\documentclass [11pt,letterpaper]{article}

\usepackage[authordate,backend=biber,bibencoding=utf8]{biblatex-chicago} 
\bibliography{bibliography}

\usepackage{trivfloat, graphicx} 
\usepackage{color}

\usepackage[group-separator={,}]{siunitx} 
\usepackage{booktabs} 
\usepackage[justification=justified,singlelinecheck=off,labelfont=bf]{caption} 
\DeclareCaptionType{example}
\usepackage[flushleft]{threeparttable} 

\usepackage{amsmath} 
\usepackage{geometry} 
\usepackage{authblk}

\title{String-based methods for tonal harmony: \\A corpus study of Haydn's string quartets}
\author{David R. W. Sears\thanks{Email: david.sears@ttu.edu}}
\affil{College of Visual \& Performing Arts, Texas Tech University}
\date{}

\begin{document}
	
\begin{flushleft}
	This is an original manuscript / preprint of a book chapter that is set to appear in the Oxford Handbook of Music \& Corpus Studies: \\
	
	\vspace{.2cm}
	Sears, David R. W (in press). ``String-based Methods for Tonal Harmony: A Corpus Study of Haydn’s String Quartets.'' In \textit{Oxford Handbook of Music \& Corpus Studies}, edited by Daniel Shanahan, John A. Burgoyne,
	and Ian Quinn. New York: Oxford University Press.\\
\end{flushleft}

{\let\newpage\relax\maketitle}

Theories of tonal harmony generally make three claims about musical organization: (1) the pitch events of tonal music group (or cohere) into discrete (primarily tertian) sonorities; (2) the succession of these sonorities over time follows a logical order, what has commonly been called \textit{harmonic syntax}; and (3) the stability relations characterizing these sonorities apply recursively, such that a triad at one level of the hierarchy---say, for example, the tonic---nests (or subsumes) sonorities at lower levels---the dominant or predominant. Thus, like language, tonal music exhibits certain \textit{design features}---namely, recurrence, syntax, and recursion---that both exploit and reflect the sensory and cognitive mechanisms by which listeners organize sensory stimuli \autocite{Fitch2006}. As a result, allusions to principles of linguistic organization abound in music research \autocite{Lerdahl1983, Rohrmeier2011a}. \textcite{Patel2008} has argued, for example, that ``the vast majority of the world's music is syntactic, meaning that one can identify both perceptually discrete elements ... and norms for the combination of these elements into sequences" \autocite[pp. 241-242]{Patel2008}.

Yet despite recent strides by the linguistics community to discover potentially analogous organizational principles in natural languages using data-driven methods, applications of statistical modeling procedures have yet to gain sufficient traction in music research. To be sure, \textcite{Neuwirth2013} has characterized the prevailing approach adopted by many in the music theory community as one based on what statistician David Fischer has called ``intuitive statistics'' \autocite[p. 34]{Neuwirth2013}, with scholars frequently eschewing explicit statistical methods in favor of qualitative descriptions derived from empirical observation. Thus, this chapter considers how we might adapt string-based methods from fields like corpus linguistics and natural language processing to address music-analytic questions related to the discovery of musical organization, with particular attention devoted to the analysis of tonal harmony. 

Following \textcite{Gjerdingen1988}, I begin in the next section by applying the taxonomy of mental structure first proposed by \textcite{Mandler1979} to the concept of musical organization. Using this taxonomy as a guide, I then present evidence in the next three sections for each of the design features mentioned above---recurrence, syntax, and recursion---using a corpus of Haydn string quartets. 

\section*{Musical Organization: Strings \& Schemes}
In the context of expressive communication systems like natural language or tonal music, it might help to clarify what the term `organization' really means, at least as it is intended here. In the simplest sense, musical organization refers to the relationships between events on the musical surface, be they notes, chords, motives, phrases, or any other coherent `units' of that organization. Following the classes of mental structure described by \textcite{Mandler1979}, we might represent the connections between these events as unordered (or \textit{coordinate}) relations, such as the members of a major triad, temporally ordered (or \textit{proordinate}) relations, such as the progression from dominant to tonic at the end of a phrase (i.e., V--I), or hierarchical (or \textit{superordinate}/\textit{subordinate}) relations, such as the prolongation of a given harmony through other (subordinate) harmonies (e.g., I--V$^4_3$--I$^6$). Bearing these relational types in mind, tonal harmony is thus an \textit{emergent} organizational system in that superordinate structures emerge out of the coordinate and proordinate relations among events on the surface. The real challenge here, then, is to model these relational types using strings.

Among computer scientists, a \textit{string} is a finite set of discrete symbols---a database of nucleic acid sequences, a dictionary of English words, or for the purposes of this study, a corpus of Haydn string quartets. In the first two cases, the mapping between the individual character or word in a printed text and its symbolic representation in a computer database is essentially one-to-one. Music encoding is considerably more complex. Notes, chords, phrases, and the like are characterized by a number of different features, so digital encodings of individual events must concurrently represent multiple properties of the musical surface. To that end, many symbolic formats encode standard music notation as a series of discrete event sequences (i.e., strings) in an $m \times n$ matrix, where $m$ denotes the number of events in the symbolic representation (e.g., notes as notated in a score), and $n$ refers to the number of encoded features or attributes (e.g., pitch, onset time, rhythmic duration, etc.).

To model the coordinate relations (i.e., vertical sonorities) associated with tonal harmony using unidimensional strings, corpus studies typically limit the investigation to a particular chord typology from music theory (e.g., Roman numerals, figured bass nomenclature, or pop chord symbols), and then identify chord events using either human annotators \autocite{Burgoyne2012,Declercq:2011,Tymoczko:2011}, or rule-based computational classifiers \autocite{Temperley:1999,Rowe:2001}. Yet unfortunately, existing typologies depend on a host of assumptions about the sorts of simultaneous relations the researcher should privilege (e.g., triads and seventh chords), and may also require additional information about the underlying tonal context, which again must be inferred either during transcription \autocite{Margulis2008}, or using some automatic (key-finding) method. \textcite{White2015} distinguishes this `top-down' approach from the `bottom-up', data-driven methods that build composite representations of chord events from simpler representations of note events \autocite{Cambouropoulos2015,Conklin2002,Quinn2010,Quinn2011, Sapp2007}.

With a representation scheme in place, researchers then divide the corpus into contiguous sequences of $n$ events (called \textit{n}-grams) to model the proordinate relations between harmonies. The resulting \textit{n}-gram distributions serve as input for tasks associated with pattern discovery \autocite{Conklin2002}, classification \autocite{Conklin2013}, automatic harmonic analysis \autocite{Taube1999}, and prediction \autocite{Sears2018}. And yet since much of the world's music is hierarchically organized such that certain events are more central (or prominent) than others \autocite{Bharucha1983}, non-contiguous events often serve as focal points in the sequence \autocite{Gjerdingen2014}. For this reason,  corpus studies employing string-based methods often suffer from the \textit{contiguity fallacy}---the assumption that note or chord events on the musical surface depend only on their immediate neighbors \autocite{Sears2017}.

By way of example, consider the closing measures of the main theme from the final movement of Haydn's string quartet Op. 50, No. 2, shown in Example \ref{ex:haydn_ex}a. The passage culminates in a perfect authentic cadence that features a conventional harmonic progression and a falling upper-voice melody. In the music theory classroom, students are taught to reduce this musical surface to a succession of chord symbols, such as the Roman numeral annotations shown below. Yet despite the ubiquity of these harmonies throughout the history of Western tonal music, existing string-based methods generally fail to retrieve this sequence of chords due to the presence of intervening embellishing tones (shown in gray), a limitation one study has called the \textit{interpolation problem} \autocite{Collins2014}.

\begin{example}[t!]
	\centering
\begingroup%
  \makeatletter%
  \providecommand\color[2][]{%
    \errmessage{(Inkscape) Color is used for the text in Inkscape, but the package 'color.sty' is not loaded}%
    \renewcommand\color[2][]{}%
  }%
  \providecommand\transparent[1]{%
    \errmessage{(Inkscape) Transparency is used (non-zero) for the text in Inkscape, but the package 'transparent.sty' is not loaded}%
    \renewcommand\transparent[1]{}%
  }%
  \providecommand\rotatebox[2]{#2}%
  \ifx\svgwidth\undefined%
    \setlength{\unitlength}{463.22855393bp}%
    \ifx\svgscale\undefined%
      \relax%
    \else%
      \setlength{\unitlength}{\unitlength * \real{\svgscale}}%
    \fi%
  \else%
    \setlength{\unitlength}{\svgwidth}%
  \fi%
  \global\let\svgwidth\undefined%
  \global\let\svgscale\undefined%
  \makeatother%
  \begin{picture}(1,0.32951736)%
    \put(0,0){\includegraphics[width=\unitlength,page=1]{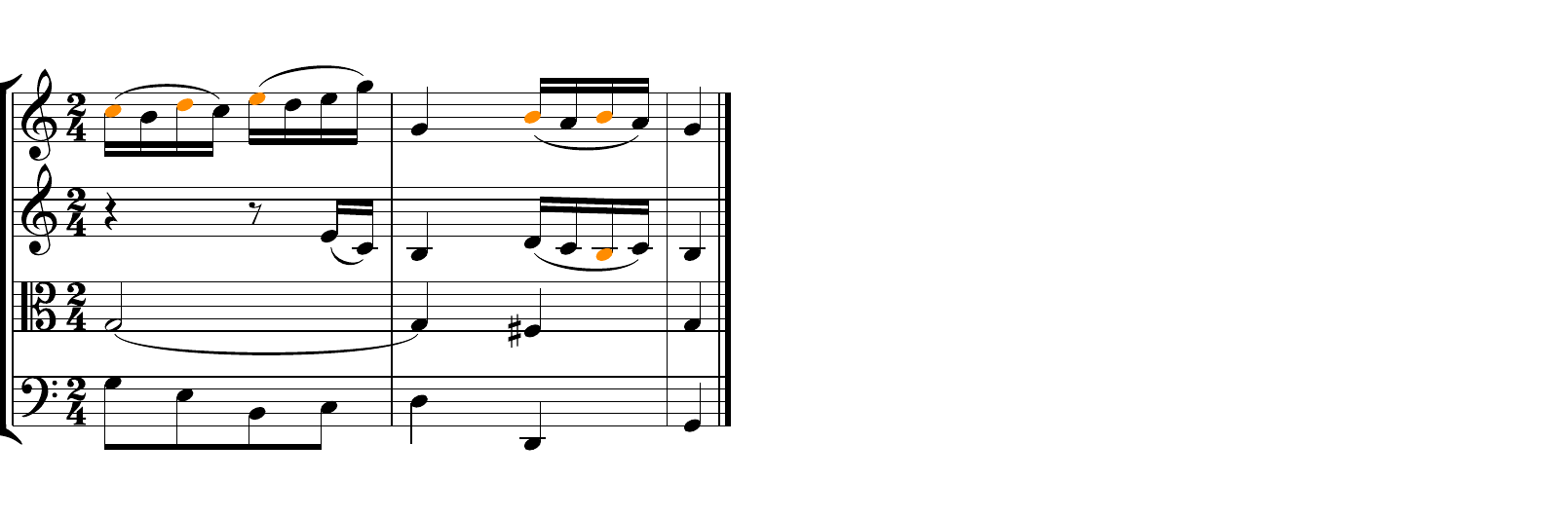}}%
    \put(0.02947566,0.31058772){\color[rgb]{0,0,0}\makebox(0,0)[lb]{\smash{(a)}}}%
    \put(0.06349122,0.00594947){\color[rgb]{0,0,0}\makebox(0,0)[lb]{\smash{I}}}%
    \put(0.10107583,0.00638702){\color[rgb]{0,0,0}\makebox(0,0)[lb]{\smash{IV$^6$}}}%
    \put(0.15543038,0.00638702){\color[rgb]{0,0,0}\makebox(0,0)[lb]{\smash{I$^6$}}}%
    \put(0.19534433,0.00599485){\color[rgb]{0,0,0}\makebox(0,0)[lb]{\smash{IV}}}%
    \put(0.2543248,0.00638702){\color[rgb]{0,0,0}\makebox(0,0)[lb]{\smash{V$^6_4$}}}%
    \put(0.32776122,0.00599485){\color[rgb]{0,0,0}\makebox(0,0)[lb]{\smash{V$^7$}}}%
    \put(0.43555133,0.00599485){\color[rgb]{0,0,0}\makebox(0,0)[lb]{\smash{I}}}%
    \put(0,0){\includegraphics[width=\unitlength,page=2]{Fig1_COLOR.pdf}}%
    \put(0.5438777,0.3106064){\color[rgb]{0,0,0}\makebox(0,0)[lb]{\smash{(b)}}}%
    \put(0.53340879,0.00484393){\color[rgb]{0,0,0}\makebox(0,0)[lb]{\smash{\small $<$0,5,$\perp$,$\perp$$>$}}}%
    \put(0.73294088,0.00450797){\color[rgb]{0,0,0}\makebox(0,0)[lb]{\smash{\small $<$7,0,4,$\perp$$>$}}}%
    \put(0.90466931,0.00450797){\color[rgb]{0,0,0}\makebox(0,0)[lb]{\smash{\small $<$0,4,$\perp$,$\perp$$>$}}}%
    \put(0,0){\includegraphics[width=\unitlength,page=3]{Fig1_COLOR.pdf}}%
  \end{picture}%
\endgroup%

	\caption{(a) Haydn, String Quartet in C major, Op. 50/2, iv, mm. 48--50. Embellishing tones are shown with gray noteheads, and Roman numeral annotations appear below. (b) Expansion. Downbeat chord onsets are annotated with the chromatic scale-degree combination (csdc) scheme for illustrative purposes.}
	\label{ex:haydn_ex}
\end{example}

Thus, the following sections consider whether string-based computational methods can discover (1) the most recurrent harmonies on the musical surface (i.e., coordinate relations); (2) the syntactic progressions that characterize a given idiom (i.e., proordinate relations); and (3) the recursive hierarchy by which certain harmonies are more central (or prominent) than others (i.e., superordinate/subordinate relations). To that end, I have developed a representation scheme that loosely approximates Roman numeral symbols using a corpus of 50 expositions from Haydn string quartets \autocite{Sears2016}. In addition to the symbolic encodings, the corpus includes accompanying text files with manual annotations for the key, mode, modulations, and pivot boundaries in each movement. Thus, I will sidestep the key-finding problem, which has already received considerable attention elsewhere (e.g., \citeauthor{Temperley2008} \citeyear{Temperley2008}). What interests me here, and consequently provides the impetus for the following pages, are the methods we use to discover the syntactic or recursive structures described in many theories of harmony. Hence, the corpus will serve as a toy dataset, with the hope that we might apply these methods to larger datasets in future work.

\section*{Coordinate Relations: Representation \& Recurrence}

Corpus studies in music research often treat the \textit{note} event as the unit of analysis, examining features like chromatic pitch \autocite{Pearce2004}, melodic interval \autocite{Vos1989}, and chromatic scale degree \autocite{Margulis2008}. Using computational methods to identify \textit{composite} events like triads and seventh chords in complex polyphonic textures is considerably more complex, since the number of distinct $n$-note combinations associated with any of the above-mentioned features is enormous. Thus, many music analysis software frameworks derive chord progressions from symbolic corpora by first performing a \textit{full expansion} of the symbolic encoding \autocite{Conklin2002}, which duplicates overlapping note events at every unique onset time.\footnote{In the Humdrum toolkit, this technique is called \textit{ditto} \autocite{Huron1993}, while \textit{Music21} calls it \textit{chordifying} \autocite{Cuthbert2010}.} Shown in Example \ref{ex:haydn_ex}b, expansion produces 14 distinct onset times. This partitioning method is admittedly too fine-grained to resemble the Roman numeral analysis in Example \ref{ex:haydn_ex}a, but provides a useful starting point for the reduction methods that follow.

To relate the chord event at each onset to an underlying tonic, some studies use the opening key signature, with the researcher determining the mode from the score, resulting in chord distributions that often fail to control for modulations or changes in modality \autocite{Margulis2008}. Key-finding algorithms have also become more common in recent decades, allowing researchers to automatically identify the key of a passage with high degrees of accuracy ($>90\%$) \autocite{Albrecht2013}. Nevertheless, the lack of available annotated corpora indicating modulations and changes of mode makes testing these algorithms quite difficult. Since in this case the corpus includes annotations for the key, mode, modulations, and pivot boundaries in each movement, we can simply map each note event to a chromatic scale degree (or \texttt{csd}) modulo 12. This scheme consists of twelve distinct symbols numbered from 0 to 11, where 0 denotes the tonic, 7 the dominant, and so on. Absent instrumental parts for each distinct onset receive an undefined symbol $\perp$.

We may now represent each chord onset as a chromatic scale-degree combination (or \texttt{csdc}) to examine the recurrence of sonorities on the musical surface. Each onset contains between one and four note events, so the initial vocabulary consists of $13^4$ (or $28,561$) possibilities. To reduce the vocabulary of possible chord types, \textcite{Quinn2010} excluded voice doublings and allowed permutations between the upper parts, so we can adopt that approach here. Thus, the major triads $\langle0, 4, 4, 7\rangle$ and $\langle0, 7, 4, 0\rangle$ would reduce to $\langle0, 4, 7, \perp\rangle$. These exclusion criteria decrease the size of the potential vocabulary to 2784 distinct types, though in this corpus the vocabulary of \texttt{csdc} consisted of just 688 types.\footnote{Ideally, we would reduce the vocabulary to less than, say, 100 symbols, but given the number of combinatorial possibilities for three- and four-note chords, I will instead introduce a novel reduction method in the final section of this chapter.}

\begin{figure}[t!]
	\centering
	\input{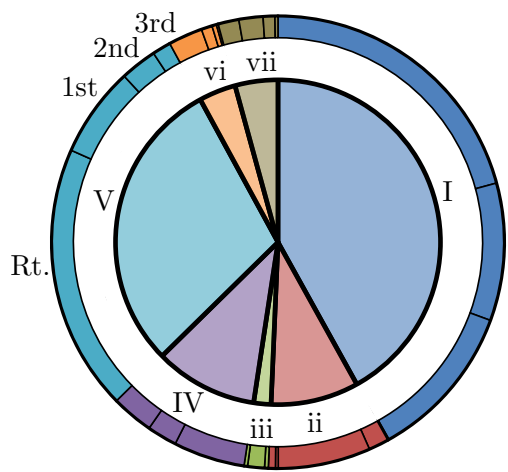}
	\caption{Multi-level pie plot of the diatonic chromatic scale degree combinations consisting of at least three chromatic scale degrees from the major mode (chord onsets located within the boundaries of a pivot were excluded). The inner pie and outer ring represent diatonic harmony (triads and seventh chords) and inversion (root position, first, second, third), respectively. Inversions appear in clockwise order for each harmony beginning in root position (labels only provided for dominant harmony). Roman numerals iii and IV did not appear in third inversion. $N = 11,253$.}
	\label{fig:csdc_pie}
\end{figure}

In total, 38\% of the $19,570$ onsets in the corpus consisted of fewer than three distinct chromatic scale degrees (e.g., $<$$0,4,\perp,\perp$$>$), so I have omitted those onsets in order to examine the most common chord types. The multi-level pie plot in Figure \ref{fig:csdc_pie} presents the onsets consisting of at least three chromatic scale degrees from major-mode passages in the corpus, with the proportions weighted by the rhythmic duration of each onset (see \textcite{Sears2016} for further details). The inner pie plot represents the diatonic harmonies with Roman numeral notation, with upper and lower case Roman numerals denoting major and minor triads, respectively. The outer concentric circle represents each inversion (root position, first, second, and third), with the inversions appearing in clockwise order for each harmony beginning in root position.

In major-mode passages, tonic harmony appeared most frequently, followed by dominant harmony, the predominant harmonies IV and ii, and finally vii, vi, and iii. In the outer concentric circle, root position chords predominated for harmonies like I, IV, V, and vi, but unsurprisingly, first inversion chords appeared more frequently for ii, iii, and vii. What is more, the 49 \texttt{csdc} types representing diatonic harmony---triads and seventh chords for every diatonic harmony in every inversion---accounted for approximately 62\% of the three- and four-note combinations in major-mode passages of the corpus.

Together, these findings suggest that (1) the most central sonorities in most theories of harmony are also the most frequent in this corpus, and (2) like their natural language counterparts, the chromatic scale-degree combinations follow a power-law distribution between frequency and rank, with the most frequent (top-ranked) types---the diatonic harmonies of the tonal system---accounting for the vast majority of the three- and four-note combinations in the corpus. Of course, these claims are by no means new. Frequency distributions of both words and chords often display power-law (or \textit{Zipfian}) distributions \autocite{Zipf1935, Rohrmeier2008, Sears2017}. What would be new is to provide evidence that the statistical regularities characterizing a tonal corpus also reflect the \textit{order} in which its constituent harmonies occur. To that end, the next section introduces string-based methods for the identification and ranking of recurrent temporal patterns. 

\section*{Proordinate Relations: Syntax \& Skip-grams}

In corpus linguistics, researchers often discover recurrent multi-word expressions (sometimes called \textit{collocations}) by dividing the corpus into sub-sequences of $n$ events (or $n$-grams), and then determining the number of instances (or \textit{tokens}) associated with each unique $n$-gram \textit{type}. \textit{N}-grams consisting of one, two, or three events are often called \textit{unigrams}, \textit{bigrams}, and \textit{trigrams}, respectively, while longer \textit{n}-grams are typically represented by the value of \textit{n}. The previous discussion represented the corpus using unigrams, for example, but to discover the most conventional (i.e., syntactic) harmonic progressions using the current representation scheme, we need only increase the value of \textit{n}.

Each piece $m$ consists of a contiguous sequence of combinations, so let $k$ represent the length of the sequence in each movement, and let $C$ denote the total number of movements in the corpus. The number of contiguous \textit{n}-gram tokens in the corpus is
\begin{equation}\label{eq:1}
\displaystyle\sum_{m=1}^{C} k_m-n+1
\end{equation}

Table \ref{tab:top10} presents the top ten contiguous bigram types ranked by count. The combinations in each type are represented with the \texttt{csdc} scheme, but for most types I also include Roman numeral annotations. In short, nine of the top ten types repeat tonic or dominant harmony, or scale degrees $\hat{1}$ or $\hat{5}$, with the top-ranked type, I--I, featuring 440 tokens. The seventh-ranked type, V$^7$--I, is in fact the only non-repeating bigram to crack the top ten. Thus, the musical surface contains considerable repetition, thereby obscuring the kinds of patterns we might hope to study (e.g., harmonic progressions containing more than one distinct harmony). Perhaps worse, recall that a significant portion of the chromatic scale degree combinations in the corpus feature fewer than three distinct chromatic scale degrees.

\begin{table}[t!]
	\centering
	\caption{Top ten contiguous bigram types ranked by count.}
	\begin{threeparttable}
		\renewcommand{\TPTminimum}{\columnwidth} 
		\makebox[\textwidth]{
			\begin{tabular}{S[table-format=2.0]ccllccrcllcc}
				\toprule
				& & \multicolumn{5}{c}{Without Exclusion Criteria} &       & \multicolumn{5}{c}{With Exclusion Criteria} \\
				\cmidrule{3-7}\cmidrule{9-13}
				\multicolumn{1}{c}{Rank} & & \textit{N} & \multicolumn{2}{c}{\texttt{csdc} (mod 12)} & \multicolumn{2}{c}{RN} &       & \textit{N} & \multicolumn{2}{c}{\texttt{csdc} (mod 12)} & \multicolumn{2}{c}{RN} \\
				\midrule
				1   &  & 440   & $0,4,7,\perp$ & $0,4,7,\perp$ & I     & I     &       & 147   & $7,2,5,11$ & $0,4,\perp,\perp$   & V$^7$    & I \\
				2    & & 212   & $7,0,4,\perp$ & $7,0,4,\perp$ & I$^6_4$   & I$^6_4$   &       & 59    & $7,0,4,\perp$ & $7,2,11,\perp$ & I$^6_4$     & V \\
				3    & & 212   & $0,4,\perp,\perp$   & $0,4,\perp,\perp$   & I     & I     &       & 56    & $11,2,5,7$ & $0,4,7,\perp$ & V$^6_5$   & I \\
				4    & & 182   & $4,0,7,\perp$ & $4,0,7,\perp$ & I$^6$    & I$^6$    &       & 52    & $0,2,5,11$ & $0,4,\perp,\perp$   & (vii) & I \\
				5    & & 154   & $0,4,\perp,\perp$   & $0,4,7,\perp$ & I     & I     &       & 42    & $7,0,4,\perp$ & $7,2,5,11$ & I$^6_4$   & V$^7$ \\
				6    & & 153   & $7,2,11,\perp$ & $7,2,11,\perp$ & V     & V     &       & 39    & $7,2,5,\perp$ & $7,0,4,\perp$ & V$^7$   & I$^6_4$ \\
				7    & & 147   & $7,2,5,11$ & $0,4,\perp,\perp$   & V$^7$    & I     &       & 31    & $7,0,4,\perp$ & $7,2,5,\perp$ & I$^6_4$    & V$^7$ \\
				8    & & 139   & $7,2,5,11$ & $7,2,5,11$ & V$^7$    & V$^7$    &       & 30    & $5,2,7,11$ & $4,0,7,\perp$ & V$^4_2$   & I$^6$ \\
				9    & & 137   & $7,\perp,\perp,\perp$     & $7,\perp,\perp,\perp$     &       &       &       & 28    & $5,2,9,\perp$ & $7,0,4,\perp$ &  ii$^6$     & I$^6_4$ \\
				10   & & 105   & $0,\perp,\perp,\perp$     & $0,\perp,\perp,\perp$     &       &       &       & 27    & $5,0,9,\perp$ & $4,0,7,\perp$ & IV   & I$^6$ \\
				\bottomrule
			\end{tabular}%
		}
		\begin{tablenotes}[flushleft]
			\small
			\item \textit{Note.} Exclusion criteria: (1) either chord contains only one distinct chromatic scale degree (\textit{monophony}); (2) neither chord contains at least three distinct chromatic scale degrees (\textit{polyphony}); (3) chords share the same chromatic scale degrees regardless of inversion (\textit{identity}); (4) chords share the same chromatic scale degree in the bass and subsets or supersets of chromatic scale degrees in the upper parts (\textit{similarity}). Parentheses suggest a change of harmony from one chord to the other, but with a pedal in the bass.
		\end{tablenotes}
	\end{threeparttable}%
	\label{tab:top10}
\end{table}

Corpus linguists typically solve problems like this by removing (or \textit{filtering}) bigram types reflecting parts of speech (or syntactic categories) ``that are rarely associated with interesting linguistic expressions'' \autocite[31]{Manning1999}. For instance, researchers often exclude types containing articles like `the' or `a' in natural language corpora to ensure that adjective-noun and noun-noun expressions will receive higher ranks in the distribution. For our purposes, one could easily extend this sort of thinking to harmonic corpora by excluding \textit{n}-grams according to the temporal periodicity or proximity of their constituent members. \textcite{Symons2012}, for example, discovered recurrent contrapuntal patterns in a corpus of two-voice solfeggi by sampling events at regular temporal intervals. Similarly, I increased the ranking of conventional cadential progressions like ii$^6$-I$^6_4$-V$^7$-I by privileging patterns with temporally proximal members \autocite{Sears2016}.\footnote{In this instance, I$^6_4$ refers to a double suspension above the cadential dominant, which is more commonly notated as V$^6_4$, as is the case in Example \ref{ex:haydn_ex}. Unfortunately, this dominant embellishment may only be determined from the immediate harmonic context (e.g., V$^{6-5}_{4-3}$ vs. I$^6_4$-I$^6$), so the present encoding scheme cannot distinguish six-four inversions of the tonic from six-four embellishments of the dominant. Thus, I have retained the I$^6_4$ annotation for instances of $<$7,0,4,$\perp$$>$ in the analyses that follow.}

There are, of course, a number of reasons to exclude patterns, but for the present study, I will exclude bigram types if they fail to represent what \textcite[231]{Meyer2000b}  referred to as ``forceful harmonic progressions": progressions featuring a genuine harmonic (i.e., pitch) change between primarily tertian sonorities.\footnote{\textcite[231]{Meyer2000b} explains, ``... the perception and cognition of patterns (and hence the formation of schemata) are dependent on clear differentiation between successive stimuli. More specifically, forceful harmonic progression depends in part on the amount of pitch change between successive triads.''} To that end, I have excluded bigram types if (1) either chord contains only one distinct chromatic scale degree (\textit{monophony}, e.g., $<$0,$\perp$,$\perp$,$\perp$$>$); (2) neither chord contains at least three distinct chromatic scale degrees (\textit{polyphony}, e.g., $<$0,$\perp$,$\perp$,$\perp$$>$$\rightarrow$$<$0,2,$\perp$,$\perp$$>$); (3) chords share the same chromatic scale degrees regardless of inversion (\textit{identity}, e.g., $<$0,4,7$\perp$$>$$\rightarrow$$<$4,0,7,$\perp$$>$); and (4) chords share the same chromatic scale degree in the bass and subsets or supersets of chromatic scale degrees in the upper parts (\textit{similarity}, e.g., $<$7,5,11,$\perp$$>$$\rightarrow$$<$7,2,5,11$>$). The first two criteria ensure that the filtered bigram types will feature tertian sonorities in some way, while the latter two criteria emphasize the importance of pitch change from one sonority to the next.

Shown in the right-most columns of Table \ref{tab:top10}, 34\% of the 5378 bigram types in the corpus met these exclusion criteria. With exclusion, progressions deemed `cadential' in most theories of harmony rose to the top of the table, with five of the top ten progressions featuring a six-four embellishment of the dominant. Along with these progressions, the table also includes typical tonic-prolongational progressions like V$^6_5$--I, V$^4_2$--I$^6$, and IV--I$^6$. The appeal of filtering in this way is thus that potentially meaningful progressions emerge out of distributional statistics. Nevertheless, by only including the counts for contiguous bigram types, we necessarily omit syntactic progressions with intervening embellishing tones from the final count. In a previous study, for example, I found that the progression I$^6$--ii$^6$--V$^7$--I \textit{never} appears contiguously in this corpus despite the apparent ubiquity of the pattern in the classical style \autocite{Sears2017}. Of the thousands of chord onsets examined here, it therefore seems unreasonable to assume that the conventional progressions in the right-most columns of Table \ref{tab:top10} should feature so few tokens. In this case, the commitment to contiguous \textit{n}-grams---the standard method in musical corpus research---has effectively tied our hands.

\begin{figure}[t!]
	\centering
	\def\svgwidth{.5\textwidth}
	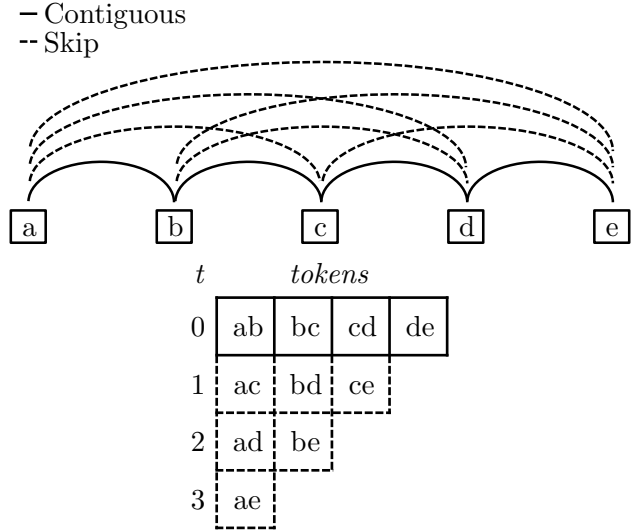
	\caption{Top: A five-event sequence, with arcs denoting all contiguous (solid) and non-contiguous (dashed) bigram tokens. Bottom: All tokens, with $t$ indicating the number of skips between events.}
	\label{fig:non_contiguous}
\end{figure}

To discover associations lying beneath (or beyond) the musical surface, we might simply relax the contiguity assumption to ensure potentially relevant bigram types appear in the distribution. Shown in Figure \ref{fig:non_contiguous}, the top plot depicts the contiguous and non-contiguous bigram tokens for a 5-event sequence with solid and dashed arcs, respectively. According to Equation \eqref{eq:1}, the number of contiguous tokens in a 5-event sequence is $k-n+1$, or four tokens. If we also include all possible non-contiguous relations, the number of tokens is given by the combination equation:
\begin{equation}
{k \choose n} = \frac{k!}{n!(k-n)!} = \frac{k(k-1)(k-2) \ldots (k-n+1)}{n!}
\end{equation}

The notation ${k \choose n}$ denotes the number of possible combinations of $n$ events from a sequence of $k$ events. By including the non-contiguous associations, the number of tokens for a 5-event sequence increases to 10. As \textit{n} and \textit{k} increase, the number of patterns can very quickly become unwieldy: a 20-event sequence, for example, contains 190 possible tokens. To overcome the combinatoric complexity of counting tokens in this way, researchers in natural language processing have limited the investigation to what I have called \textit{fixed-skip} \textit{n}-grams \autocite{Sears2017,Guthrie2006}, which only include \textit{n}-gram tokens if their constituent members occur within a fixed number of skips $t$. Shown in the bottom plot in Figure \ref{fig:non_contiguous}, $ac$ and $bd$ constitute one-skip tokens (i.e., $t=1$), while $ad$ and $be$ constitute two-skip tokens. Thus, up to 10 tokens appear in a 5-event sequence when $t=3$.

By relaxing the restrictions on the possible associations between events in a sequence, the skip-gram method has increased our chances of including potentially meaningful \textit{n}-grams in the final distribution. It does not ensure that the most frequent types will feature genuine harmonic progressions, however. To be sure, the skip-gram method aggregates the counts for bigram types whose members occur \textit{within} a certain number of skips, and so is just as susceptible to the most repetitive patterns on the surface. As a result, repeating bigram types like I--I tend to retain their approximate ranking regardless of the permitted number of skips.

To resolve this issue, we could again exclude patterns that do not represent a genuine harmonic change between adjacent members, but corpus linguists also frequently employ alternative ranking functions whose goal is to discover recurrent multi-word expressions using other relevance criteria. In this case, the skip-gram method provides counts for each \textit{n}-gram type at a number of possible skips. If the harmonies in conventional harmonic progressions tend to appear in strong metric positions and feature intervening embellishing tones, we might instead rank patterns using a statistic that characterizes the \textit{depth} at which conventional harmonic progressions tend to emerge.

\begin{figure}[t!]
	\centering
	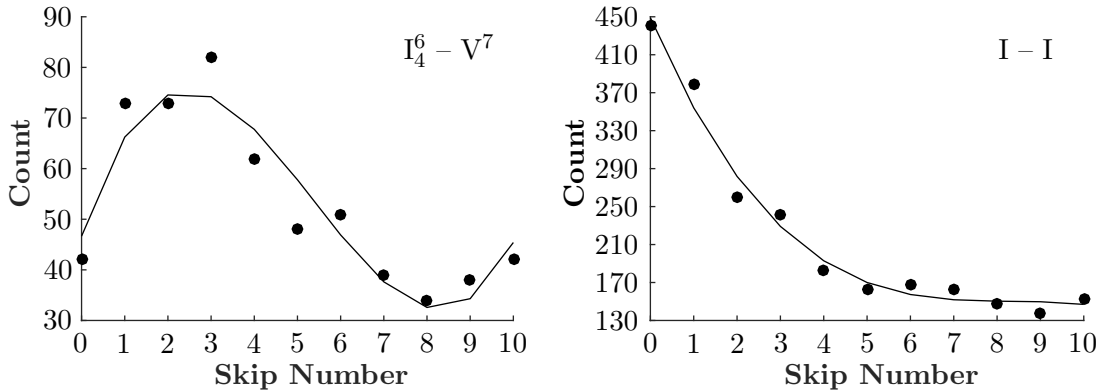
	\caption{Scatter plots of counts at each skip for the highest- and lowest-ranked bigram types: $<$7,0,4,$\perp$$>$ $\rightarrow$ $<$7,2,5,11$>$ (left) and $<$0,4,7,$\perp$$>$ $\rightarrow$ $<$0,4,7,$\perp$$>$ (right). Counts in both plots are fit with a third-order polynomial trend (solid line).}
	\label{fig:poly_plot}
\end{figure}

Figure \ref{fig:poly_plot} presents scatter plots of the counts measured from zero to ten skips for the cadential six-four progression, I$^6_4$--V$^7$, and the top-ranked type in the corpus, I--I. Recall that I--I features 440 tokens on the surface (see Table \ref{tab:top10}). As the number of skips $t$ between the members of this type increase, the associated counts decrease. In other words, repetitive patterns like I--I appear far more prevalently at (or near) the surface. This result seems unsurprising---presumably over enough skips, \textit{all} bigram types become less frequent. Yet since conventional harmonic progressions often appear in strong metric positions and feature intervening embellishing tones, we might assume that the counts for these patterns should actually \textit{increase} with $t$ up to a certain point, and then decrease for more distal relations. This is in fact exactly what we find for the I$^6_4$--V$^7$ progression in Figure \ref{fig:poly_plot}, which features its highest count when $t=3$, but decreasing counts when $t>3$.

So how might we privilege patterns like I$^6_4$--V$^7$ in the final ranking? If conventional harmonic progressions tend to appear beneath the surface, the counts across skips should increase as $t$ increases. We could then model this assumption by fitting a first-order polynomial (or \textit{linear}) trend to the distribution of counts for each $n$-gram type. The best-fit line modeled by the equation $y_i = \beta_1x_i +\beta_0$ minimizes the error between the predicted count at each skip and its actual count in Cartesian space (called \textit{linear regression}), where the leading coefficient $\beta_1$ characterizes the shape of the trend (i.e., the slope of the line). Thus, positive estimates of $\beta_1$ would indicate an increasing trend, whereas negative estimates would indicate a decreasing trend.

In this model, ranking bigram types using $\beta_1$ privileges patterns whose counts increase as $t$ increases. As a result, conventional progressions like I--V$^7$ and V--I appear in the top ten, but so do syntactic retrogressions like ii$^6$--I (table not shown). In this case, permitting such large skips ensures that the final distribution will features types whose members skip over syntactically meaningful harmonies \textit{within} the progression. Thus, we could revise the linear model by assuming that the counts should increase as $t$ initially increases (e.g., up to $t=3$), and then \textit{decrease} for larger skips. A second-order polynomial trend---which would produce a U-shape if the leading coefficient $\beta_2$ is positive, and an inverted U-shape if $\beta_2$ is negative---would peak near the center of the distribution (i.e., at around $t=5$), so I have ranked each bigram type using the leading coefficient $\beta_3$ of a third-order polynomial trend. Patterns that increase from zero to approximately three skips, and then decrease for larger numbers of skips, will produce the positive trend found for the I$^6_4$--V$^7$ progression in Figure \ref{fig:poly_plot}. Conversely, patterns that decrease exponentially from zero to ten skips will produce the negative trend found for the I--I progression. In point of fact, these two patterns are the highest- and lowest-ranked types in the distribution.

Table \ref{tab:top10poly} presents the top ten bigram types ranked by $\beta_3$. Even without exclusion criteria, the top ten types include several conventional harmonic progressions, such as V$^7$--I, a pre-dominant-to-I$^6_4$ progression, and a tonic-prolongational progression from I to IV$^6_4$. With exclusion criteria, a few other interesting patterns emerge, such as the progression from I$^6$ to IV, or the tonic pedal supporting a progression from vii to I. Perhaps more importantly, few of the patterns discovered using this method appear frequently on the surface. The sixth-ranked V$^7$--I progression, for example, includes just three tokens on the surface. Even across the top one hundred patterns in the distribution, the median count is just four tokens. Thus, it seems that patterns appearing just beneath the surface---i.e., whose counts peak at around $t=3$---feature many of the conventional (or syntactic) progressions described in most theories of harmony.

Yet despite the success of this ranking function to privilege patterns representing a genuine harmonic change of some sort, a persistent problem remains: namely, four of the top ten progressions in Table \ref{tab:top10poly} are variants of the same V$^7$--I progression. This finding suggests that the vocabulary, at 688 symbols, is simply too large to serve as a suitable proxy for the chord vocabularies in theories of harmony. Discovering the most recurrent, syntactic progressions---or reducing the musical surface to a sequence of its most central (or salient) harmonies---requires a novel vocabulary reduction method, a problem I turn to in the next section.

\section*{Superordinate/Subordinate Relations: Recursion \& Reduction}

The appeal of the scheme selected for this study is that the most common chromatic scale degree combinations will have analogues in most theories of harmony. Frankly, this is not surprising given that the scheme relies on human annotations about the keys and modes for each movement. Nevertheless, the \texttt{csdc} representation is more promiscuous than traditional definitions of `chord' would embrace. Whereas theorists tend to assign chordal status only to those vertical sonorities featuring stacked intervals of a third, the \texttt{csdc} scheme makes no distinctions between chord tones and non-chord tones, consonant and dissonant intervals, or diatonic and chromatic scale degrees \autocite{Quinn2010}. Hence, the vocabulary of \texttt{csdc} types is enormous.

\begin{table}[t!]
	\centering
	\caption{Top ten bigram types ranked by leading coefficient of third-order polynomial trend.}
	\begin{threeparttable}
		\renewcommand{\TPTminimum}{\columnwidth} 
		\makebox[\textwidth]{
			\begin{tabular}{S[table-format=2.0]ccllccrcllcc}
				\toprule
				& & \multicolumn{5}{c}{Without Exclusion Criteria} &       & \multicolumn{5}{c}{With Exclusion Criteria} \\
				\cmidrule{3-7}\cmidrule{9-13}
				\multicolumn{1}{c}{Rank} & & \textit{$\beta_3$} & \multicolumn{2}{c}{\texttt{csdc} (mod 12)} & \multicolumn{2}{c}{RN} &       & \textit{$\beta_3$} & \multicolumn{2}{c}{\texttt{csdc} (mod 12)} & \multicolumn{2}{c}{RN} \\
				\midrule
				1   &  & .429  & $7,0,4,\perp$ & $7,2,5,11$ & I$^6_4$   & V$^7$    &       & .429  & $7,0,4,\perp$ & $7,2,5,11$ & I$^6_4$   & V$^7$ \\
				2    & & .240  & $5,\perp,\perp,\perp$     & $0,\perp,\perp,\perp$ & &    &       & .181  & $0,4,7,\perp$   & $0,5,9,\perp$ & I     & IV$^6_4$ \\
				3    & & .220  & $7,\perp,\perp,\perp$     & $2,\perp,\perp,\perp$ & &    &       & .171  & $7,5,11,\perp$ & $0,4,\perp,\perp$     & V$^7$    &  I  \\
				4    & & .181  & $0,4,7,\perp$   & $0,5,9,\perp$ & I     & IV$^6_4$    &       & .166  & $5,9,\perp,\perp$ & $7,0,4,\perp$ & PrD    & I$^6_4$ \\
				5    & & .173  & $2,\perp,\perp,\perp$     & $7,\perp,\perp,\perp$     &       &       &       & .162  & $0,4\perp,\perp$   & $0,5,9,\perp$ & I     & IV$^6_4$ \\
				6    & & .171  & $7,5,11,\perp$ & $0,4,\perp,\perp$     & V$^7$    &  I        &       & .155  &  $7,2,5,11$ & $0,4,\perp,\perp$     & V$^7$    &  I \\
				7    & & .166  & $5,9,\perp,\perp$ & $7,0,4,\perp$ & PrD    & I$^6_4$    &       & .144  &  $7,2,5,11$ & $0,4,7,\perp$     & V$^7$    &  I \\
				8    & & .162  & $0,4\perp,\perp$   & $0,5,9,\perp$ & I     & IV$^6_4$   &       & .141  & $4,0,7,\perp$   & $5,0,9,\perp$ & I$^6$     & IV \\
				9    & & .161  & $0,4,7,\perp$& $0,\perp,\perp,\perp$     & I    &       &       & .135  & $7,2,5,\perp$ & $0,4,\perp,\perp$ & V$^7$    &  I \\
				10   & & .155  & $7,2,5,11$ & $0,4,\perp,\perp$     & V$^7$    &  I     &       & .122  & $0,2,5,11$ & $0,4,7,\perp$ & (vii)    &  I \\
				\bottomrule
			\end{tabular}%
		}
		\begin{tablenotes}[flushleft]
			\small
			\item \textit{Note.} A third-order polynomial trend fit to the counts from zero to ten skips for each bigram type (i.e., $y_i = \beta_3x^3_i + \beta_2x^2_i + \beta_1x_i +\beta_0$). PrD denotes predominant function. See Table \ref{tab:top10} for exclusion criteria.
		\end{tablenotes}
	\end{threeparttable}%
	\label{tab:top10poly}
\end{table}

How, then, do we solve the harmonic reduction problem when the relations between note events explode in combinatorial complexity for complex polyphonic textures? To demonstrate that these sorts of organizational systems emerge out of distributional statistics, let us consider again the two ranking functions from the previous section: (1) \textit{count}, which assumes that the most relevant or meaningful types are the most frequent; and (2) a \textit{polynomial trend}, which assumes that the most relevant types tend to appear just beneath the surface (e.g., at $t=3$). The former is a simple statistic that represents the sum of the instances for each type, whereas the latter exploits domain-specific knowledge about the corpus under investigation. There are limitations to both ranking functions, however. One could argue, for example, that since the top-ranked types in Table \ref{tab:top10poly} are by no means the most frequent, the assumption that they are somehow relevant (or conventional) is unwarranted. Similarly, corpus linguists have argued that count is not a sufficient indicator for a strong attraction between words \autocite[5]{Evert2008}, since two highly frequent words are also likely to co-occur quite often just by chance. V$^7$ and I appear frequently in the corpus (see Figure \ref{fig:csdc_pie}), for example, so it is possible that their appearance in Table \ref{tab:top10} simply reflects the joint probability of their co-occurrence.

To resolve this issue, corpus linguists have developed a large family of \textit{association} (or \textit{attraction}) \textit{measures} that quantify the statistical association between co-occurring words \autocite{Evert2008}. The majority of these measures use contingency tables. Table \ref{tab:contingency} presents the contingency table for the most common bigram type associated with the V$^7$--I progression: $<$7,2,5,11$>$ $\rightarrow$ $<$0,4,$\perp$,$\perp>$. The counts reflect tokens with up to five skips between bigram members. The table has four cells for tokens containing both chord$_1$ and chord$_2$ (a), tokens containing chord$_1$ but not chord$_2$ (i.e., any other chord) (b), tokens containing chord$_2$ but not chord$_1$ (c), and tokens containing neither chord (d). The \textit{marginal frequencies}, so called because they appear at the margins of the table, represent the sum of each row and column. Thus, the co-occurence frequency for V$^7$--I is 581.

Again, there are dozens of available association measures \autocite{Pecina2009}, but \textit{Fisher's exact test} is perhaps the most appropriate (or mathematically rigorous) significance test for the analysis of contingency tables \autocite{Agresti2002}. The mathematical formalism need not concern us here, but in short, Fisher's exact test computes the total probability of all possible outcomes that are similar to or more extreme than the observed contingency table. The resulting probability (or \textit{p-value}) will be large if the two chords of a given bigram are statistically independent, but very small if the two chords are unlikely to co-occur at the estimated frequency just by chance. In this case, the \textit{p}-value is vanishingly small ($p < .0001$), suggesting that V$^7$ and I are statistically dependent.

Essentially, association measures produce empirical statements about the \textit{statistical attraction} between chords. They do not measure the potential asymmetry of this association, however. This is to say that in many cases chord$_1$ could be more (or less) predictive of chord$_2$ than the other way around, so \textcite{Gries2013} and \textcite{Nelson2014} have suggested alternative association measures based on the predictive asymmetry between the members of each bigram. According to \textcite{Firth1957}, association measures should quantify the statistical influence an event exerts on its neighborhood, where some events exert more influence than others. Given such a measure, we could reduce the chord vocabulary by privileging chords that exert the greatest `attractional force' on their neighbors.

Asymmetric (or \textit{directional}) association measures typically compute the conditional probabilities between the members of each bigram \autocite{Michelbacher2007}. In Table \ref{tab:contingency}, for example, the probability that I follows V$^7$ can be computed from the frequencies in the first row. In this case, $P(\text{I}|\text{V}^7)$ is $\frac{a}{a+b}$, with $a$ representing all of the instances in which I follows V$^7$, and $b$ representing all of the instances in which some other harmony follows V$^7$. Thus, $\frac{581}{3873}=.15$, which tells us that I follows V$^7$ roughly 15\% of the time. This estimate does not represent the probability that V$^7$ \textit{precedes} I, however. To compute this statistic, we can use the frequencies in the first column of Table \ref{tab:contingency}. Here, $P(\text{V}^7|\text{I})$ is $\frac{a}{a+c}$, or $\frac{581}{6216}=.09$. Thus, for this particular variant of the V$^7$--I progression, V$^7$ is a better predictor of I than the other way around. Or put another way, I exerts the greater attractional force.

\begin{table}[t!]
	\centering
	\caption{Contingency table for the bigram $<$7,2,5,11$>$ $\rightarrow$ $<$0,4,$\perp$,$\perp>$ (V$^7$--I). Counts reflect tokens with up to five skips between chord events.}
	\setlength{\tabcolsep}{16pt}
	\def\arraystretch{1.5}
	\begin{tabular}{r|S[table-format=4.0] S[table-format=6.0]r|S[table-format=6.0]}
		& {I}     & {Not I}    &       & Totals \\
		\midrule
		V$^7$    & 581 \enspace {(a)}   & 3292 \enspace {(b)}  &       & 3873 \\
		Not V$^7$   & 5635 \enspace {(c)}  & 106862 \enspace {(d)} &       & 112497 \\
		\midrule
		Totals & 6216  & 110154 &       & 116370 \\
	\end{tabular}%
	\label{tab:contingency}%
\end{table}%

I have formalized this statistical inference in the following way:
\begin{equation}
\textsc{asym} = P(\text{chord}_2|\text{chord}_1) - P(\text{chord}_1|\text{chord}_2) = \frac{a}{a+b} - \frac{a}{a+c}
\end{equation}
In this equation, \textsc{asym} is simply the arithmetic difference between the two conditional probabilities. The estimates of \textsc{asym} fall in a range between $-1$ and 1, where positive values indicate that chord$_2$ is the attractor, negative values indicate that chord$_1$ is the attractor, and 0 indicates bidirectionality, since both harmonies exert equivalent attractional force. For the bigram in Table \ref{tab:contingency}, the positive directional asymmetry of .06 tells us that I exerts more influence on V$^7$, and so serves as the statistical attractor within the bigram.

It bears mentioning here that directional asymmetries differ from the \textit{temporal} asymmetries described in theories of harmony. Whereas temporal asymmetries refer to the conventionality or syntactic plausibility of harmonies according to their temporal order (e.g., ii$^6$--V$^7$ compared to V$^7$--ii$^6$), directional asymmetries attempt to capture the attractional force between two harmonies in a \textit{specified} temporal relationship. Thus, V$^7$--ii$^6$ may be much less likely---and thus, less syntactic---than ii$^6$--V$^7$, but our notion of directional asymmetry simply indicates which of the two harmonies is the stronger attractor in both progressions.

To reduce the vocabulary of chord types, we could simply calculate the number of bigram types in which each unigram type is the attractor. Shown in Table \ref{tab:asym_ranks}, \textit{N}$_{\text{attractor}}$ indicates that $<$$0,4,7,\perp$$>$ serves as the attractor in 604 distinct bigram types in the corpus. Unsurprisingly, \%$_{\text{attractor}}$ also tells us that this variant of I is the attractor for every bigram in which it appears. Finally, the table also presents an alternative asymmetric measure based on the sum of the asymmetries for each bigram type in which the indicated unigram type is a member. In this case, I exerts the greatest attractional force of any of unigram type in the corpus $\sum_{\textsc{asym}}=56.41$, with other harmonies like V, V$^7$, I$^6$, and ii$^6$ also making the top ten. Vocabulary reduction methods could then use a table like this one to create an \textit{$n$-best list}, which uses a specified threshold $n$ to determine the members (and non-members) of the vocabulary \autocite{Evert2008}. We would then assimilate harmonies appearing below this threshold into those appearing above using a kind of incremental clustering method.

A discussion of clustering methods for directional asymmetry data deserves its own study, but for the sake of illustration, I have presented one possible method in Example \ref{ex:reduction}. In this case, reducing the musical surface to a sequence of its most central harmonies is a specific case of the more general vocabulary reduction problem considered thus far. Starting with the attractional force rankings represented in Table \ref{tab:asym_ranks}, a simple harmonic reduction algorithm could reduce a sequence of harmonies---and thus, the size of the overall vocabulary---by linking the chord exerting the least attractional force in the sequence, denoted by $c_i$, to the left or right chord neighbor exerting the greater attractional force ($c_{i-1}$ or $c_{i+1}$). The algorithm would then remove $c_i$ from the sequence and repeat the process until all of the chords have been linked.

Shown in Example \ref{ex:reduction}, the harmonic reduction algorithm just described could be used to produce a tree diagram not unlike those found in Lerdahl and Jackendoff's (1983) approach. In this case, the chord in onset three, $<$$9,0,7,\perp$$>$, features the least attractional force of the chords in the sequence, so the algorithm linked onset three to the stronger attractor, which in this case was the right neighbor, $<$$9,0,5,\perp$$>$ (i.e., IV$^6$). The algorithm then removed onset three from the sequence and started the process again, at each step linking the combination of scale degrees exerting the least attractional force to the stronger adjacent attractor, and then removing that combination from the sequence. Branches reaching above the horizontal dashed line produce the reduction shown in the bottom system. Thus, for this passage the algorithm pruned all of the chords containing embellishing tones, resulting in the sequence of chords corresponding to the Roman numeral annotations provided below.

\begin{table}[t!]
	\centering
	\caption{Top ten unigram types, ranked according to the number of bigram types in which each unigram type is the attractor.}
	\begin{threeparttable}
		\renewcommand{\TPTminimum}{\columnwidth} 
		\makebox[\textwidth]{
			\begin{tabular}{S[table-format=2.0]S[table-format=3.0]S[table-format=3.1]clc}
				\toprule
				\multicolumn{1}{c}{Rank} & \textit{N}$_{\text{attractor}}$ & \%$_{\text{attractor}}$ & $\sum_{\textsc{asym}}$ &  \multicolumn{1}{l}{\texttt{csdc} (mod 12)} & RN \\
				\midrule
				1     & 604   & 100   & 56.41 &  $0,4,7,\perp$ & I \\
				2     & 560   & 99.6  & 38.33 &  $0,4,\perp,\perp$   & I \\
				3     & 555   & 99.3  & 31.35 &  $7,\perp,\perp,\perp$     &  \\
				4     & 509   & 98.1  & 36.38 &  $7,2,11,\perp$ & V \\
				5     & 508   & 98.5  & 29.10  &  $7,2,5,11$ & V$^7$ \\
				6     & 508   & 97.7  & 23.89 &  $0,\perp,\perp,\perp$     &  \\
				7     & 497   & 98.8  & 35.34 &  $4,0,7,\perp$ & I$^6$ \\
				8     & 440   & 96.9  & 32.27 &  $7,0,4,\perp$ & I$^6_4$ \\
				9     & 425   & 95.6  & 27.36 &  $5,2,9,\perp$ & ii$^6$ \\
				10   & 412 & 96.3   & 13.98  & $2,\perp,\perp,\perp$ & \\
				\bottomrule
			\end{tabular}%
		}
		\begin{tablenotes}[flushleft]
			\small
			\item \textit{Note.} \textit{N}$_{\text{attractor}}$ denotes the number of bigram types in which each unigram type is the attractor, and \%$_{\text{attractor}}$ indicates the percentage of bigram types in which each unigram type appears as the attractor.
		\end{tablenotes}
	\end{threeparttable}%
	\label{tab:asym_ranks}%
\end{table}%

Presumably the pronounced directional asymmetries in the distribution of counts for each bigram type allow this algorithm to distinguish genuine harmonies from chords containing embellishing tones. Nevertheless, a toy example like this one tends to paper over the cracks of what is in fact a very difficult problem. Note, for example, that the highest branches of the tree only partly reflect how an analyst might parse this particular passage. The three harmonies at the very top of the tree seem reasonable enough (I--V$^7$--I), but this algorithm identified the next most important harmony as I$^6$ rather than IV, producing the progression, I--I$^6$--V$^7$--I. If Meyer's (\citeyear{Meyer2000b}) preference for genuine harmonic change is reasonable, then IV would be the more fitting member in the final progression even though I$^6$ obtained the higher rank in Table \ref{tab:asym_ranks}.

We could perhaps solve this problem by applying the algorithm not just to the passage in question, but to the entire corpus of movements simultaneously. At each step, the algorithm would assimilate variants of harmonies like $<$$9,0,7,\perp$$>$  into more stable attractors like $<$$9,0,5,\perp$$>$, and then adjust the ranks in Table \ref{tab:asym_ranks} before starting the process again. In this way, the attractional force for each harmony in the distribution would change from one hierarchical level to the next. One could imagine that by incrementally assimilating variants into their more stable attractors, the resulting vocabulary might better reflect the functional categories described in theories of harmony (e.g., tonic, predominant, and dominant), and so produce greater attractional force estimates for harmonies like IV relative to those like I$^6$ at higher levels of the hierarchy.

\begin{example}[t!]
	\centering
\begingroup%
  \makeatletter%
  \providecommand\color[2][]{%
    \errmessage{(Inkscape) Color is used for the text in Inkscape, but the package 'color.sty' is not loaded}%
    \renewcommand\color[2][]{}%
  }%
  \providecommand\transparent[1]{%
    \errmessage{(Inkscape) Transparency is used (non-zero) for the text in Inkscape, but the package 'transparent.sty' is not loaded}%
    \renewcommand\transparent[1]{}%
  }%
  \providecommand\rotatebox[2]{#2}%
  \ifx\svgwidth\undefined%
    \setlength{\unitlength}{215.99999246bp}%
    \ifx\svgscale\undefined%
      \relax%
    \else%
      \setlength{\unitlength}{\unitlength * \real{\svgscale}}%
    \fi%
  \else%
    \setlength{\unitlength}{\svgwidth}%
  \fi%
  \global\let\svgwidth\undefined%
  \global\let\svgscale\undefined%
  \makeatother%
  \begin{picture}(1,1.18975609)%
    \put(0,0){\includegraphics[width=\unitlength,page=1]{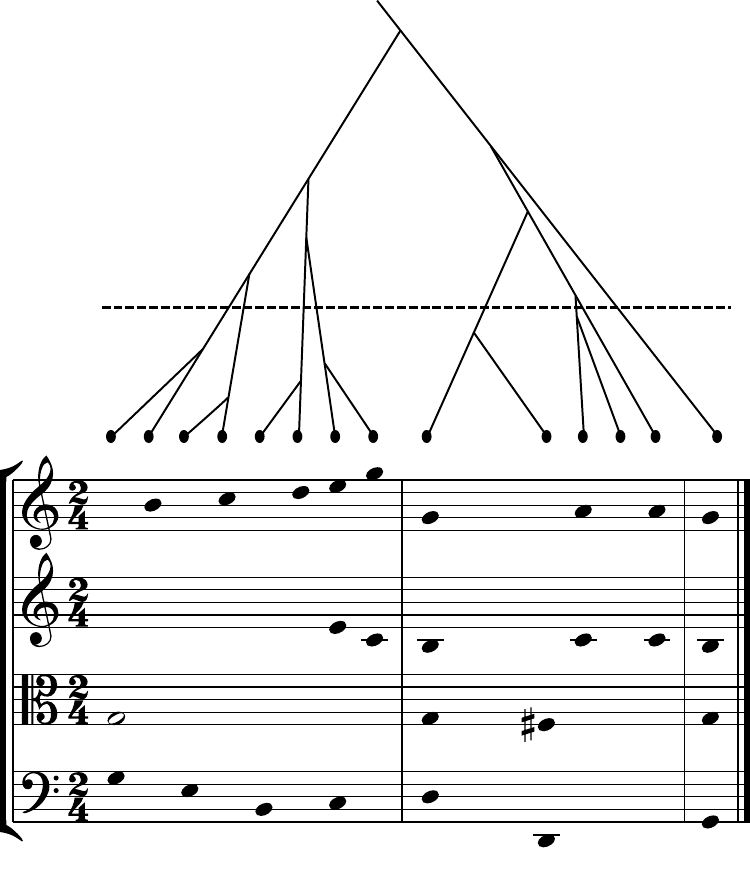}}%
    \put(0.13820203,0){\color[rgb]{0,0,0}\makebox(0,0)[lb]{\smash{I}}}%
    \put(0.21139772,0.00093836){\color[rgb]{0,0,0}\makebox(0,0)[lb]{\smash{IV$^6$}}}%
    \put(0.32796519,0.00093836){\color[rgb]{0,0,0}\makebox(0,0)[lb]{\smash{I$^6$}}}%
    \put(0.42097114,0.00009732){\color[rgb]{0,0,0}\makebox(0,0)[lb]{\smash{IV}}}%
    \put(0.54745925,0.00093836){\color[rgb]{0,0,0}\makebox(0,0)[lb]{\smash{V$^6_4$}}}%
    \put(0.70494932,0.00009732){\color[rgb]{0,0,0}\makebox(0,0)[lb]{\smash{V$^7$}}}%
    \put(0.93188387,0.00009732){\color[rgb]{0,0,0}\makebox(0,0)[lb]{\smash{I}}}%
  \end{picture}%
\endgroup%

	\caption{Tree diagram produced by the harmonic reduction algorithm for Op. 50/2, iv, mm. 48--50. Branches reaching above the horizontal dashed line produce the reduction shown in the bottom system.}
	\label{ex:reduction}
\end{example}

Despite these limitations, the important point here is that when the vocabulary is larger than, say, 5 to 10 symbols, the admittedly simple algorithm just described produces surface-to-mid-level parsings not unlike those found in a Roman numeral analysis. Thus, it seems reasonable to suggest that the statistical associations between events near the surface reflect many of the organizational principles captured by most theories of harmony.

\section*{Conclusions}

This chapter adapted string-based methods from corpus linguistics to examine the coordinate, proordinate, and superordinate/subordinate relations characterizing tonal harmony. In doing so, I have assumed that three of the organizational principles associated with natural languages---recurrence, syntax, and recursion---might also appear in tonal music. To that end, I began by examining the distribution of chromatic scale degree combinations across a corpus of Haydn string quartets. Unsurprisingly, the diatonic harmonies of the tonal system accounted for the vast majority of the combinations in the corpus. To model progressions of these harmonies over time, I then employed skip-grams, which include sub-sequences in an \textit{n}-gram distribution if their constituent members occur within a certain number of skips. After applying filtering measures and ranking functions of various types, the most relevant (or meaningful) harmonic progressions emerged at the top of the \textit{n}-gram distribution. Finally, to reduce the musical surface in Example \ref{ex:haydn_ex}a to a sequence of its most central harmonies, I presented a simple harmonic reduction algorithm (and tree diagram) based on an asymmetric probabilistic measure of attractional force. In this case, the algorithm pruned all of the chords containing embellishing tones.

To examine the potential of string-based methods in corpus studies of music, I made certain simplifying assumptions about the principles mentioned above. Perhaps the most obvious of these was to restrict the purview of coordinate relations to temporally coincident scale degrees. This restriction seems reasonable for homorhythmic textures, but much less so for string quartets, piano sonatas, and the like, which often feature accompanimental textures that prolong harmonies over time (e.g., an Alberti bass pattern). This problem was at least partly resolved by the harmonic reduction algorithm, which assimilates variants into nearby attractors (e.g., I$^6$ into I), but note that it cannot replace variants of a given harmonic category---say, for example, $<$$0,4,\perp,\perp$$>$---if their more central (or prototypical) attractors---$<$$0,4,7,\perp$$>$---do not appear nearby. Thus, the current algorithm would sometimes fail to produce convincing harmonic reductions for passages featuring complex polyphonic textures.\footnote{For an innovative string-based solution to this problem, see \textcite{White2013b}.}

Nevertheless, because these methods are ambivalent about the organizational systems they model, one need only revise the simplifying assumptions above to suit the needs of the research program. Indeed, since these methods proceed from the bottom up, we could just as easily use skip-grams or attraction measures to study melody, rhythm, or meter. Similarly, corpus studies in cognitive linguistics or systematic musicology might argue that the syntactic properties of natural language or tonal music should reflect limitations of human auditory processing, so it seems reasonable to impose similar restrictions on the sorts of contiguous and non-contiguous relations the skip-gram method should model \autocite{Sears2017}. This claim seems especially relevant if we assume that the recursive hierarchy described throughout this chapter is nonuniform and discontinuous \autocite{Meyer1973}, in that the statistics operating at relatively surface levels of musical organization---the syntax of harmonic progressions (e.g., I--ii$^6$--V--I)---might differ from those operating at deeper levels---long-range key relationships (e.g., I--III--V--I).

There are, indeed, many possible solutions for the computational problems described here. My goal was not to offer definitive results, but to demonstrate that the most recent methods developed in corpus linguistics and natural language processing have much to offer for corpus studies of music. Indeed, if \textcite{Patel2008} is right that language and music share certain fundamental design features, then skip-grams, contingency tables, and association measures represent invaluable tools for the study of tonal harmony.

\pagebreak
\printbibliography
\end{document}